\documentclass[sigconf,nonacm]{acmart}

\usepackage{booktabs}
\usepackage{pifont}

\newcommand{\fhat}{\hat{f}}
\newcommand{\xq}{x_{\mathrm{q}}}
\newcommand{\xp}{x_{+}}

\newcommand{\phimap}{\varphi}

\newcommand{\ActCoachAware}{0.572}
\newcommand{\ActCoachFiltered}{0.447}

\newcommand{\ActCoachRetained}{58}
\newcommand{\ActFaceFiltered}{0.432}
\newcommand{\ActFaceFree}{0.973}
\newcommand{\ActFaceRetained}{44}
\newcommand{\ActNiceFiltered}{0.409}
\newcommand{\ActNiceFree}{0.993}
\newcommand{\ActNiceRetained}{41}
\newcommand{\AuditCoverageMax}{1.0000}
\newcommand{\AuditCoverageMin}{1.0000}
\newcommand{\AuditMaxErr}{6.2\times 10^{-15}}

\newcommand{\AuditQueries}{2,060}
\newcommand{\CoachCoverageMin}{1.000}
\newcommand{\CoachValidityMax}{1.000}
\newcommand{\CoachValidityMin}{0.977}
\newcommand{\ControlValidityMin}{0.965}
\newcommand{\FullValidityMin}{0.967}
\newcommand{\HelocCoachManifold}{1.83}
\newcommand{\HelocCoachSparsity}{58}
\newcommand{\HelocManifoldGain}{22}
\newcommand{\HelocNiceManifold}{2.36}
\newcommand{\HelocNiceSparsity}{14}

\newcommand{\KValidityMaxEight}{1.000}

\newcommand{\KValidityMaxThree}{1.000}

\newcommand{\KValidityMinEight}{0.972}

\newcommand{\KValidityMinThree}{0.703}

\newcommand{\NSplits}{20}

\newcommand{\ReferenceMax}{1.96}
\newcommand{\ReferenceMin}{0.76}
\newcommand{\RuntimeMax}{11.4}
\newcommand{\RuntimeMin}{1.7}

\newcommand{\TaiwanCoachSparsity}{52}
\newcommand{\TaiwanManifoldGain}{39}

\newcommand{\TaiwanNiceSparsity}{7}

\newcommand{\ExQueryProba}{0.151}
\newcommand{\ExCompProba}{0.733}

\begin{document}

\title{Leaf Values as Coordinates: Exact Contrastive Explanation\\
for Gradient-Boosted Ensembles}

\author{Emanuele Luzio}
\email{emanuele.luzio@gmail.com}
\affiliation{\institution{Independent Researcher}\country{}}

\begin{abstract}
A gradient-boosted ensemble predicts by summing one leaf value per tree.
Read those values as coordinates rather than as intermediate results,
and every instance becomes a point in $\mathbb{R}^{M}$ on which the
model acts \emph{linearly}: the score is the sum of the coordinates.

This small change of view makes contrastive explanation exact. The
difference between two instances is a vector that is identically zero
wherever they share a leaf, so the gap between a rejected applicant and
an accepted one is carried by a handful of coordinates, each traceable
to a real split in a real tree. Nothing is fitted, sampled, or assumed
additive in features --- the additivity is already there, in the right
space.

We build a recourse method on this representation and evaluate it on
five tabular datasets under repeated cross-validation. Its
recommendation reconstructs the model's own decision to
$\AuditMaxErr$, so an auditor can re-check the arithmetic without the
model. On the credit datasets it is Pareto-non-dominated on effort
against realism. And when recommendations are restricted to changes the
subject could actually make --- not their age, not a settled
delinquency --- it retains $\ActCoachRetained\,\%$ of its validity where
the strongest baseline retains $\ActNiceRetained\,\%$, a distinction the
standard evaluation cannot see because it never asks whether a
recommendation can be carried out.
\end{abstract}

\maketitle

\section{Introduction}
\label{sec:intro}

Gradient-boosted ensembles are additive by construction. With $M$ trees,
\begin{equation}
  \fhat(x) \;=\; \sum_{m=1}^{M} v_m\bigl(\ell_m(x)\bigr),
  \label{eq:model}
\end{equation}
where $\ell_m(x)$ is the leaf that $x$ reaches in tree $m$ and
$v_m(\cdot)$ its value. This is normally read as an implementation
detail: the model computes $M$ numbers and adds them.

Read instead as a \emph{representation}, it says something stronger.
Define
\begin{equation}
  \phimap(x) \;=\; \bigl(v_1(\ell_1(x)),\,\dots,\,v_M(\ell_M(x))\bigr)
  \;\in\; \mathbb{R}^{M}.
  \label{eq:phi}
\end{equation}
Then $\fhat(x) = \mathbf{1}^{\top}\phimap(x)$. The model, which is
violently non-linear in the input features, is \emph{linear} in
$\phimap$ --- indeed it is the simplest possible linear functional, an
unweighted sum. All the non-linearity has been pushed into the map
$\phimap$ itself, where it is piecewise constant and exactly known.

Three consequences follow, and they are the paper.

\textbf{Differences are sparse.} For two instances $x,y$, the vector
$\phimap(x)-\phimap(y)$ is identically zero in every coordinate where
they reach the same leaf. Their score gap is therefore carried by the
trees where they diverge, and by nothing else. This is exact, not
approximate.

\textbf{Explanation is subtraction.} Reading off the non-zero
coordinates of that difference gives a complete account of why the model
scores two instances differently, in units that sum to the gap. No
surrogate is fitted, no reference distribution chosen, no assumption
made that the model is additive in \emph{features} --- which it is not,
and which is the silent assumption behind attribution-driven
explanation.

\textbf{Recourse is retrieval.} If explanation is a difference vector,
then producing recourse means choosing what to subtract from: an
accepted instance whose difference from the query is small, decisive,
and reachable. The method reduces to a choice of neighbour in
$\phimap$-space, plus the accounting that comes free with it.

We develop this into COACH, a recourse method for tabular
gradient-boosted models (Section~\ref{sec:method}), and evaluate it on
five datasets (Sections~\ref{sec:setup}--\ref{sec:results}). The
representation's central promise --- that the accounting is exact --- is
verified directly: across $\AuditQueries$ queries the reported
coordinates reconstruct the model's margin gap to $\AuditMaxErr$.

We also report a finding that is not about our method. Standard recourse
evaluation asks the model whether a modified profile would be approved,
never whether the subject could reach it. Restricting recommendations to
feasible changes costs every method we test 40--60\,\% of its validity
and reverses their ranking (Section~\ref{sec:feasible}). Methods that
change few features fare worst, which is the opposite of what the
conventional metrics reward.

\section{The Representation}
\label{sec:repr}

Write $\Delta(x,y) = \phimap(x)-\phimap(y)$ and
$\mathcal{G}(x,y) = \{m : \ell_m(x) \neq \ell_m(y)\}$ for the trees
where $x$ and $y$ diverge. By construction $\Delta(x,y)_m = 0$ for
$m \notin \mathcal{G}(x,y)$, so
\begin{equation}
  \fhat(y) - \fhat(x)
  \;=\; \sum_{m \in \mathcal{G}(x,y)}
        \bigl[v_m(\ell_m(y)) - v_m(\ell_m(x))\bigr].
  \label{eq:decomp}
\end{equation}

Equation~\ref{eq:decomp} is an identity. It is worth dwelling on how
little it asks for: no linearity in features, no independence, no
sampling, no locality. Two instances, one model, exact arithmetic.

\paragraph{From coordinates to features.}
The coordinates of $\Delta$ are indexed by trees, and a subject cannot
act on a tree. Each diverging tree is therefore attributed to the
feature at its \emph{decisive split} --- the node where the two
root-to-leaf paths first separate --- and coordinates sharing a feature
are summed. This produces a table with one row per feature: the split
threshold, the number of trees involved, and the summed contribution.

That table is simultaneously the recommendation and its justification.
Because the attribution partitions $\mathcal{G}$, the rows still sum to
the exact gap; Section~\ref{sec:exact} confirms they do at scale.

\begin{table*}[t]
\centering
\begin{center}\small
\begin{tabular}{lrrrrl}
\toprule
Feature & Query & Comp. & Threshold & $\Delta v$ (trees) & Solo $\Delta\fhat$ \\
\midrule
\texttt{NetFractionRevolvingBurden} & 81 & 0 & 31.5 & +2.260 (75) & +1.110 \\
\texttt{AverageMInFile} & 38 & 74 & 53.1 & +0.182 (14) & +0.291 \\
\texttt{MSinceOldestTradeOpen} & 85 & 160 & 120.5 & +0.197 (16) & +0.081 \\
\texttt{NumInstallTradesWBalance} & 1 & 2 & 2.0 & +0.050 (2) & +0.029 \\
\texttt{NetFractionInstallBurden} & --- & 84 & 74.6 & +0.045 (11) & +0.013 \\
\midrule
\multicolumn{4}{l}{\textit{sum of entries}} & $\mathbf{+2.734}$ & \\
\multicolumn{4}{l}{\textit{model's margin gap}} & $+2.734$ & \\
\bottomrule
\end{tabular}
\end{center}

\caption{A recommendation for a rejected HELOC applicant, scored
$\ExQueryProba$ against a threshold of $0.50$, against a comparator at
$\ExCompProba$. Each row is a feature reached by summing the coordinates
of $\Delta$ whose decisive split falls on it; \emph{trees} counts those
coordinates. The last two rows are the point of the representation: the
entries reproduce the model's own margin gap exactly, so the arithmetic
can be re-checked without the model. \emph{Solo} is a different quantity
--- the shift from moving that feature alone --- and the fact that the two
columns disagree is the feature-interaction effect that
Section~\ref{sec:method} introduces $\varepsilon$ to absorb. Note also
rows two and three: raising an account-age feature is not something an
applicant can do, which Section~\ref{sec:feasible} takes up.}
\label{tab:example}
\end{table*}

\paragraph{Similarity in this space.}
Two instances are close in $\phimap$ when they share many leaves. This
is the leaf co-occurrence long used as a proximity measure in tree
ensembles \citep{marmerola2020counterfactual}, with one difference that
matters here: because our coordinates are the leaf \emph{values} rather
than co-occurrence indicators, they carry sign and magnitude and sum to
the prediction. Co-occurrence tells you two instances are similar;
$\phimap$ tells you how their difference produces the decision.

\section{COACH}
\label{sec:method}

Given a rejected query $\xq$, COACH selects an accepted comparator
$\xp$ and returns the feature table of Section~\ref{sec:repr}.

\paragraph{Eligibility.}
Comparators are drawn from training instances the \emph{model} accepts,
not those labelled positive: a recommendation must be something the
model would actually approve. A margin $\varepsilon$ requires
$\fhat(\xp)$ to clear the threshold with surplus, which absorbs a
mismatch the representation does not remove --- the model is linear in
$\phimap$, but a subject acts on features, and moving one feature
changes leaf assignments in trees whose decisive split lies elsewhere.

\paragraph{Ranking.}
Among eligible comparators, COACH prefers those whose difference is both
small in support and concentrated in impact:
\begin{equation}
  \lambda(\xq,\xp) \;=\;
  \underbrace{\Bigl(1 - \tfrac{|\mathcal{G}|}{M}\Bigr)}_{\text{agreement}}
  \cdot
  \frac{\sum_{m \in \mathcal{G}} \lvert \Delta_m \rvert}
       {\sum_{m} \lvert \phimap(\xq)_m \rvert},
  \label{eq:leverage}
\end{equation}
divided by $1 + \beta\,d(\xq,\xp)$ with $d$ the mean
standard-deviation-normalised $L_1$ distance in feature space, so that
nearby comparators are preferred without changing who is eligible.
Section~\ref{sec:whatmatters} reports how much this ranking actually
buys, and the answer is: less than its prominence here suggests.

\paragraph{Tiers.}
Eligible comparators are split into terciles of their own score
distribution, giving the subject a progression rather than one target.
Terciles rather than fixed probability bands: a confident model on an
imbalanced problem can leave a fixed band such as $[0.60,0.65)$ holding
one instance or none, at which point every query silently receives the
same comparator, or none at all.

\paragraph{Feasibility.}
\label{sec:feas-method}
Features differ in what a subject can do with them, and
Section~\ref{sec:feasible} shows this dominates everything else. Given
labels marking each feature \textsc{mutable}, \textsc{increase-only},
\textsc{decrease-only} or \textsc{immutable}, let $F(\xq,\xp)$ be the
coordinates whose move the labels permit and $s_j$ the scale of feature
$j$. Each candidate is weighted by
\begin{equation}
  w(\xq,\xp) =
  \frac{\sum_{j \in F} \lvert x_{+,j}-x_{\mathrm{q},j}\rvert/s_j}
       {\sum_{j} \lvert x_{+,j}-x_{\mathrm{q},j}\rvert/s_j}
  \in [0,1],
  \label{eq:feasible}
\end{equation}
and the comparator returned maximises
$\lambda \cdot w / (1+\beta d)$. A candidate whose gap rests on the
applicant's age scores $w \approx 0$ before any recommendation exists.
Rows the subject cannot act on are additionally dropped from the
recommendation, while remaining in the audit trail --- those trees do
carry part of the gap, and removing them would break
Equation~\ref{eq:decomp}.

\section{Experimental Setup}
\label{sec:setup}

Five tabular datasets: FICO HELOC \citep{fico2018heloc}, Taiwan Credit
Default \citep{yeh2009taiwan}, Adult Income \citep{kohavi1996adult}, and
Polish Bankruptcy at 1- and 2-year horizons \citep{zieba2016polish}.
Models are XGBoost \citep{chen2016xgboost}, 300 trees of depth 4.
Baselines are NICE \citep{brughmans2023nice}, FACE
\citep{poyiadzi2020face}, Feature Tweaking
\citep{tolomei2017tweaking}, and greedy recourse driven by SHAP
\citep{lundberg2017shap} and LIME \citep{ribeiro2016lime}.

\emph{Validity} is the fraction of \emph{all} queries whose recourse the
model accepts, so coverage cannot be traded against it. \emph{Sparsity}
is the fraction of features changed; \emph{manifold distance} is the
normalised $L_2$ distance to the five nearest accepted training
instances. For the last we also report where \emph{genuine} accepted
instances sit relative to their own neighbours,
$\ReferenceMin$--$\ReferenceMax\,\sigma$, without which a figure in
$\sigma$ cannot be judged.

Results are $\NSplits$ repeated cross-validation splits. With five
splits the smallest attainable two-sided Wilcoxon $p$ is $0.0625$ ---
above $\alpha$ before any correction --- so a 5-fold design cannot report
significance whatever the effect size. Because repeated splits share
training data \citep{dietterich1998approximate}, we use the
Nadeau--Bengio corrected resampled $t$-test \citep{nadeau2003inference}
with Holm correction \citep{holm1979simple}, and follow
\citet{demsar2006statistical} in treating the interval on the paired
difference as primary.

\section{Results}
\label{sec:results}

\subsection{The accounting is exact}
\label{sec:exact}

The representation's central claim is that the reported rows reconstruct
the model's decision. We tested it adversarially: an independent
verifier receives the recommendation table and nothing else --- no model,
no training data --- and must recover the margin gap.

Across $\AuditQueries$ queries on five datasets, the rows account for a
fraction $\AuditCoverageMin$--$\AuditCoverageMax$ of the gap, with
reconstruction error never exceeding $\AuditMaxErr$ in margin units.
This is floating-point noise, and it is what Equation~\ref{eq:decomp}
predicts. The practical consequence is that a compliance reviewer can
verify the arithmetic linking a recommendation to a decision without
access to the model that made it.

\subsection{Recourse quality}
\label{sec:quality}

\begin{figure*}[t]
\centering
\includegraphics[width=0.92\textwidth]{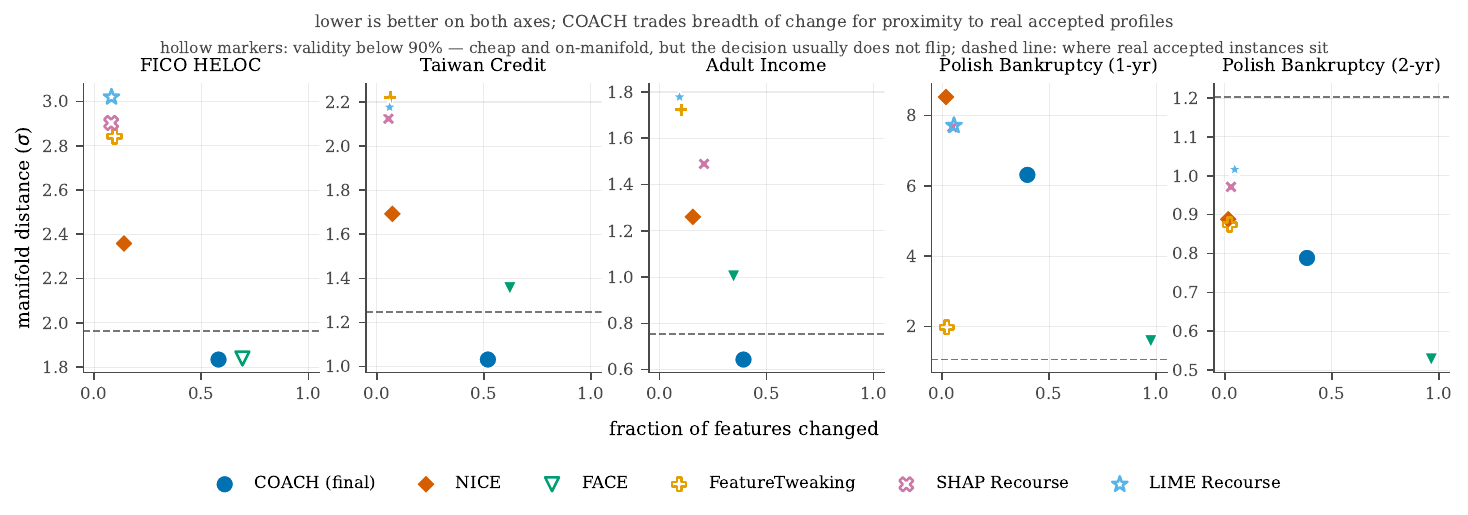}
\Description{Five scatter plots, one per dataset, with the fraction of
features changed on the horizontal axis and manifold distance in standard
deviations on the vertical axis. In each panel COACH sits toward the lower
right --- changing many features but landing close to real accepted instances
--- while NICE sits toward the upper left, changing few features but landing
further away. A dashed horizontal line marks the distance separating genuine
accepted instances from their own neighbours; COACH falls at or below it on
three of the five datasets.}
\caption{Effort against realism, five datasets. COACH occupies the
low-manifold, high-sparsity corner; NICE the opposite. Hollow markers
mark validity below 90\,\%. The dashed line is where \emph{genuine}
accepted instances sit relative to their own neighbours --- a recourse at
or below it is, by this measure, indistinguishable from a real accepted
person.}
\label{fig:tradeoff}
\end{figure*}

COACH reaches coverage $\CoachCoverageMin$ and validity between
$\CoachValidityMin$ and $\CoachValidityMax$. On HELOC it lands
$\HelocManifoldGain\,\%$ closer to real accepted instances than NICE
($\HelocCoachManifold$ vs.\ $\HelocNiceManifold$) while changing
$\HelocCoachSparsity\,\%$ of features against NICE's
$\HelocNiceSparsity\,\%$; on Taiwan, $\TaiwanManifoldGain\,\%$ closer at
$\TaiwanCoachSparsity\,\%$ versus $\TaiwanNiceSparsity\,\%$.

That trade is the honest summary. Treating validity as a precondition
and asking which methods are non-dominated on effort against realism,
COACH and NICE both survive on all five datasets, at opposite ends of
one frontier; every other baseline is dominated or fails the validity
gate somewhere. No weighting-free argument selects between the two ---
which is why we report dominance rather than a composite score, and why
the case for the representation rests on Section~\ref{sec:exact} rather
than on winning a metric.

\subsection{What the ranking is worth}
\label{sec:whatmatters}

Equation~\ref{eq:leverage} is the one part of the method that looks like
a design choice rather than a consequence of the representation. It
earns less than it appears to.

Replacing it with an \emph{arbitrary} eligible comparator --- ignoring
the query entirely --- leaves validity essentially unchanged
($\ge\ControlValidityMin$ against $\ge\FullValidityMin$). Applying a
recommendation copies the comparator's values on every decisive feature,
so the result lands on or near an accepted instance whichever one it is.
Validity comes from the eligibility constraint, not from the ranking;
$\lambda$ buys \emph{narrower} recommendations, and proximity to the
data manifold comes from the $\beta$ penalty.

We report this because it is what the representation predicts. If the
space is the right one, the operation on top of it should be simple, and
a ranking heuristic should not be doing the heavy lifting. It is not.

\subsection{Recourse the subject can act on}
\label{sec:feasible}

\begin{table}[t]
\centering\small
\caption{Validity when the subject may change only what they can change. Immutable features (age, protected attributes, settled history) are frozen and one-directional features may move only the feasible way; the same projection is applied to every method. \emph{Filtered} removes infeasible moves after retrieval; \emph{constraint-aware} folds feasibility into comparator selection. Polish Bankruptcy is excluded: we have no defensible mutability labels for 64 derived accounting ratios.}
\label{tab:actionable}
\begin{tabular}{lccc}
\toprule
Method & FICO HELOC & Taiwan Credit & Adult Income \\
\midrule
\multicolumn{4}{l}{\textit{unconstrained}} \\[2pt]
COACH & 0.967 & 1.000 & 1.000 \\
NICE & 0.978 & 1.000 & 1.000 \\
FACE & 0.922 & 0.997 & 1.000 \\
FeatureTweaking & 0.782 & 1.000 & 1.000 \\
\midrule
\multicolumn{4}{l}{\textit{feasible changes only}} \\[2pt]
COACH (constraint-aware) & 0.648 & 0.315 & 0.752 \\
COACH (filtered) & 0.547 & 0.087 & 0.708 \\
NICE (filtered) & 0.372 & 0.115 & 0.740 \\
FACE (filtered) & 0.343 & 0.187 & 0.767 \\
FeatureTweaking (filtered) & 0.208 & 0.090 & 0.893 \\
\bottomrule
\end{tabular}
\end{table}

Every validity figure above --- ours and, as far as we can tell,
everyone's --- is scored by a model with no concept of what a person can
do. The worked example in Section~\ref{sec:repr} is typical: alongside
``pay down your revolving balance'' it asks the applicant to raise their
average account age from 38 to 74 months, that is, to have opened their
accounts three years earlier than they did.

Labelling features by mutability and restoring the original value
wherever a recommended move is impossible, we re-score every method
under the same projection. Table~\ref{tab:actionable} gives the result.
NICE falls from $\ActNiceFree$ to $\ActNiceFiltered$, FACE from
$\ActFaceFree$ to $\ActFaceFiltered$. Three points follow.

\emph{The cost is large and common to all methods}, which locates it in
the evaluation protocol rather than in any algorithm --- it is a
correction to published numbers we did not produce.

\emph{The ordering reverses.} NICE leads unconstrained and trails
constrained. A practitioner choosing on published validity would pick
the method that degrades worst.

\emph{Sparsity is the property that breaks.} A recommendation touching
two or three features has no slack: if one is immutable, most of it is
deleted and the remainder rarely moves the decision. A broad
recommendation degrades gracefully. COACH retains
$\ActCoachRetained\,\%$ of its validity against $\ActNiceRetained\,\%$
for NICE and $\ActFaceRetained\,\%$ for FACE --- for the same reason the
conventional evaluation penalises it.

Finally, \emph{where} the constraint is applied matters. Filtering
infeasible moves after retrieval gives $\ActCoachFiltered$; folding
feasibility into selection via Equation~\ref{eq:feasible} gives
$\ActCoachAware$. That gap exceeds every between-method difference in
the unconstrained comparison. We could implement it for one method only
--- it needs a retrieval step to modify --- so we offer it as a
demonstration rather than a general law.

\subsection{Further checks}
\label{sec:further}

\paragraph{How much of a broad recommendation must be acted on.}
Section~\ref{sec:feasible} treats breadth as protective; the fair
objection is that a thirteen-feature recommendation is not actionable
either. Acting on only the $k$ highest-impact rows retains validity
between $\KValidityMinThree$ and $\KValidityMaxThree$ at $k{=}3$, and
between $\KValidityMinEight$ and $\KValidityMaxEight$ at $k{=}8$.
Breadth is not an all-or-nothing demand --- which is also why it has
slack to lose.

\paragraph{Other libraries and deeper trees.}
Equation~\ref{eq:phi} is a property of the leaf structure, not of an
implementation, so it should transfer. Re-running the protocol under
LightGBM \citep{ke2017lightgbm} gives validity within a point of
XGBoost on every dataset. Across
$\{$XGBoost, LightGBM$\}\times\{4,6,8\}$ depths $\times\{300,800\}$
trees, leaf co-occurrence falls as expected --- on HELOC mean agreement
drops from about $0.51$ at depth 4 to $0.35$ at depth 8 --- but
recommendations do not lengthen and validity is unaffected. The
agreement term of Equation~\ref{eq:leverage} is a ranking signal;
degrading it changes which comparator wins, not whether the winner
works.

\paragraph{Cost.}
Retrieval takes $\RuntimeMin$--$\RuntimeMax$\,ms per query over a
16$\times$ range of pool sizes. It scales with the number of
\emph{eligible} comparators rather than the pool, so it grows fastest on
datasets where the model accepts most of the population. The comparison
itself is a vectorised operation over a leaf-index matrix; a deployment
with a far larger pool would want approximate nearest neighbours on leaf
fingerprints, which we have not implemented.

\section{Related Work}
\label{sec:related}

\paragraph{Tree representations.}
Using leaf structure as a similarity space is not new: random-forest
proximities date to \citet{breiman2001random}, and
\citet{marmerola2020counterfactual} use leaf co-occurrence for
counterfactual search. Those constructions are \emph{indicator}-valued
--- they record whether two instances share a leaf. Equation~\ref{eq:phi}
keeps the leaf \emph{values}, so the coordinates are signed, carry
magnitude, and sum to the model's output. That is what turns a
similarity measure into an exact decomposition of a decision, and it is
the distinction on which this paper rests.

\paragraph{Tree-aware recourse.}
\citet{tolomei2017tweaking} tweak features toward the nearest
positive-prediction leaf; \citet{cui2015action} extract optimal actions
by integer programming; \citet{parmentier2021ocean} give a
mixed-integer formulation with manifold constraints; and
\citet{lucic2022focus} relax the ensemble into a differentiable
surrogate. All search the tree structure for a point. We retrieve an
existing accepted instance and report the exact accounting; an optimiser
can find a cheaper point than any instance in the reference population,
whereas retrieval makes the justification the selection mechanism
itself.

\paragraph{Model-agnostic recourse and its evaluation.}
Optimisation methods \citep{wachter2017counterfactual,mothilal2020dice}
treat the classifier as an oracle. Attribution-driven recourse
\citep{lundberg2017shap,ribeiro2016lime} assumes a per-feature score
predicts the effect of moving that feature; for a model additive in
trees but not in features this is false, and silently so. Instance-based
methods \citep{poyiadzi2020face,brughmans2023nice} borrow real accepted
examples, keeping recommendations grounded without explaining them.
\citet{ustun2019recourse} and \citet{karimi2021algorithmic} constrain
\emph{generation} by actionability and causality;
Section~\ref{sec:feasible} instead quantifies what omitting the
constraint costs at \emph{evaluation} time, for the many methods that
make no feasibility claim. \citet{pawelczyk2021carla} standardise
recourse benchmarking, and the projection used here is the kind of check
such a framework could adopt directly.

\section{Limitations}
\label{sec:limitations}

The mutability labels are judgements about the world, not facts in the
data. Ours are documented per feature; where unsure we marked a feature
mutable, which constrains less and makes the reported costs lower
bounds. We have no defensible labels for Polish Bankruptcy's 64 derived
accounting ratios, so the constrained analysis covers three datasets.

Constrained recourse is measured, not solved: COACH survives the
constraint better than the baselines and still lands at
$\ActCoachAware$, which nobody should deploy. Getting further needs
retrieval that \emph{searches} for feasible comparators rather than
down-weighting infeasible ones.

The audit trail is verified, not validated. We show the artifact is
exact and independently checkable; we have not shown that a compliance
officer benefits from it, and until a study exists that claim is a
design argument.

Finally, the representation is $M$-dimensional and tied to one trained
model. It is an interpretation device, not a transferable embedding: a
retrained ensemble induces a different $\phimap$.

\section{Conclusion}
\label{sec:conclusion}

Treating leaf values as coordinates rather than intermediate results
makes a gradient-boosted model linear in a space we can write down
exactly. Contrastive explanation then reduces to subtraction, and the
result is exact by construction rather than by approximation --- verified
here to $\AuditMaxErr$ over $\AuditQueries$ queries.

Building recourse on that representation gives a method competitive on
the conventional metrics and, more usefully, one that degrades
gracefully when recommendations are restricted to changes a person can
actually make. That last comparison is one the standard evaluation
cannot make at all, and it reverses the field's ranking when it is made.

\bibliographystyle{ACM-Reference-Format}
\bibliography{coach_refs}

\end{document}